%% file: main.tex
\documentclass{article}

\usepackage{PRIMEarxiv}

\usepackage[utf8]{inputenc} 
\usepackage[T1]{fontenc}    
\usepackage{url}            
\usepackage{booktabs}       
\usepackage{amsfonts}       
\usepackage{nicefrac}       
\usepackage{microtype}      
\usepackage{lipsum}
\usepackage{fancyhdr}       
\usepackage{graphicx}       
\graphicspath{{media/}}     

\usepackage{multirow}
\usepackage{array}
\usepackage{xcolor}
\usepackage{booktabs}
\usepackage{amssymb}

\title{Lived Experience Matters: \\Automatic Detection of Stigma on Social Media toward People Who Use Substances}

\author{
  Salvatore Giorgi \\
  National Institute on Drug Abuse \\
  sal.giorgi@nih.gov \\ 
   \And
  Douglas Bellew \\
  National Institute on Drug Abuse \\
   \And
  Daniel Roy Sadek Habib \\
  National Institute on Drug Abuse \\
   \AND
  Garrick Sherman \\
  National Institute on Drug Abuse \\
   \And
  Jo\~{a}o Sedoc \\
  New York University \\
   \And
  Chase Smitterberg \\
  National Institute on Drug Abuse \\
   \AND
  Amanda Devoto \\
  National Institute on Drug Abuse \\
   \And
   McKenzie Himelein-Wachowiak\\
  National Institute on Drug Abuse \\
   \And
  Brenda Curtis \\
  National Institute on Drug Abuse \\
  brenda.curtis@nih.gov \\
}

\begin{document}
\maketitle

\begin{abstract}
Stigma toward people who use substances (PWUS) is a leading barrier to seeking treatment.
Further, those in treatment are more likely to drop out if they experience higher levels of stigmatization. 
While related concepts of hate speech and toxicity, including those targeted toward vulnerable populations, have been the focus of automatic content moderation research, stigma and, in particular, people who use substances have not. 
This paper explores stigma toward PWUS using a data set of roughly 5,000 public Reddit posts.
We performed a crowd-sourced annotation task where workers are asked to annotate each post for the presence of stigma toward PWUS and answer a series of questions related to their experiences with substance use.
Results show that workers who use substances or know someone with a substance use disorder are more likely to rate a post as stigmatizing.
Building on this, we use a supervised machine learning framework that centers workers with lived substance use experience to label each Reddit post as stigmatizing. 
Modeling person-level demographics in addition to comment-level language results in a classification accuracy (as measured by AUC) of 0.69 -- a 17\% increase over modeling language alone. 
Finally, we explore the linguist cues which distinguish stigmatizing content: PWUS substances and those who don't agree that language around othering (``people'', ``they'') and terms like ``addict'' are stigmatizing, while PWUS (as opposed to those who do not) find discussions around specific substances more stigmatizing.
Our findings offer insights into the nature of perceived stigma in substance use. 
Additionally, these results further establish the subjective nature of such machine learning tasks, highlighting the need for understanding their social contexts. 
\end{abstract}


\input{01intro}

\input{02relatedwork}

\input{03data}

\input{04annotationtask}
\input{05methods}
\input{06results}
\input{07discussion}

\input{08conclusions}

\section*{Acknowledgments}
We thank Mitchell L. Gordon for sharing his jury learning code. This research was supported in part by the Intramural Research Program of the NIH, National Institute on Drug Abuse (NIDA).

\bibliographystyle{unsrt}  
\bibliography{references}  

\appendix

\input{supplement}

\end{document}

%% file: 01intro.tex
\section{Introduction}
\label{sec: intro}

In the U.S. in 2021, 61.2 million people aged 12 or older (22\% of the population) used substances, and 46.3 million (17\% of the population) met the criteria for having a substance use disorder (SUD)~\cite{substance2021}. 
Despite the prevalence of SUDs and substance use, 94\% of people with a SUD did not receive treatment.
There are significant barriers to seeking treatment, including stigma (negative biases including stereotypes, prejudice, and discrimination~; \cite{corrigan2002understanding}).
Studies have shown that, of those people who did perceive a need for treatment, 22.7\% report stigma as a reason for lack of seeking out treatment~\cite{ashford2019abusing}.
Stigma has measurable consequences on the health and well-being of people who use substances (PWUS): it contributes to diminished help-seeking~\cite{clement2015impact}, medication non-adherence~\cite{sirey2001stigma}, healthcare avoidance~\cite{byrne2008healthcare}, worse healthcare~\cite{van2013stigma}, poor health outcomes~\cite{byrne2008healthcare,stangl2019health}, and lower quality of life~\cite{cheng2019negative}. 
Indeed, stigma, in general, is a central driving force for population mortality~\cite{hatzenbuehler2013stigma}.  

While extensive research has been conducted on the automatic detection of related concepts of hate speech and toxic language on social media~\cite{fortuna2018survey}, stigma and, in particular, stigma towards PWUS has received relatively little attention. 
This is despite the fact that roughly half of the people in treatment for SUDs have reported that their online communities contain triggering content~\cite{ashford2018technology}.
The standard pipeline for the automatic detection of hate speech on social media is to collect a corpus of posts (from Twitter or Reddit, for example) and label each post as to whether or not it contains hate speech through a crowd-sourcing annotation task. Typically produced through a majority vote across the annotations, these labels are then used to train a machine learning classifier to detect hate speech on unseen data automatically. 
Recently, several issues have been identified with this pipeline where the annotation process and majority voting introduce substantial biases into the final machine learning model through, for example, annotator demographics~\cite{diaz2018addressing}, annotator beliefs~\cite{sap-etal-2022-annotators}, insensitivity to dialects of minority populations~\cite{sap-etal-2019-risk}
Thus, increasing focus is being given to understanding \emph{who} is annotating data, what are the annotators' beliefs, moral values, and lived experiences, and how can machine learning methods incorporate dissenting opinions and disagreement~\cite{davani-etal-2022-dealing,prabhakaran2021releasing,rottger-etal-2022-two}. See Uma et al. (2021)~\cite{uma2021learning} for a survey and in-depth discussion on disagreement in learning tasks. 

In this paper, we attempt to automatically identify stigmatizing content on social media by centering people who are the subject of the stigma -- those who have lived experience with substance use. This is done in an attempt to understand both manifestations of stigma and \emph{who} perceives it through three research questions:
\begin{itemize}
\item \textbf{RQ1} Can stigma towards people who use substances (PWUS) be automatically identified?
\item \textbf{RQ2} Does lived experience with substance use inform how stigma is perceived?
\item \textbf{RQ3} Are there linguistic differences between stigma perceived by people with lived experience with substance use and those without?
\end{itemize}
To do this, we collect a sample of 5,000 Reddit comments that contain mentions of substances or substance use and run a crowd-sourcing task where we pay Amazon Mechanical Turk (Mturk) workers to label the posts as having stigma towards PWUS. We also ask the workers a series of demographic and substance use-related questions. Using this demographic information, we assign stigma labels to each Reddit comment through a learning framework that allows one to center various demographic distributions, also known as jury learning~\cite{gordon-jury-learning}. We end by examining the linguistic cues associated with stigma across different populations.


\paragraph{Contributions} Our key contributions include: (1) the public release of a data set of stigma-annotated Reddit comments along with demographic variables of the annotators; (2) we show that annotators with lived experience with substance use are more likely to label a social media post as stigmatizing through the evaluation of a machine learning classifier which centers groups of annotators with common attributes; and (3) we identify linguistic markers associated with stigmatizing social media posts as highlighted by annotators with lived experience with substance use.

%% file: 02relatedwork.tex
\section{Related Work}
\label{sec: related work}

\subsubsection{Definition of Stigma Toward People Who Use Substances}
Stigma can be thought of as a collection of negative biases against certain groups of people, which often incorporate three components: stereotypes, prejudice, and discrimination~\cite{corrigan2002understanding}. 
All three components can manifest through interpersonal interactions or intrapersonally, known as self-stigma.
Following Link and Phelan (2021)~\cite{link2001conceptualizing}, stigma consists of the ``identification of differentness, the construction of stereotypes, the separation of labeled persons into distinct categories, and the full execution of disapproval, rejection, exclusion, and discrimination'' by people with access to ``social, economic, and political power''. 

While many group experience stigma, this paper is focused on stigma experienced by people who use substances or people with a SUD.
In a population of people in treatment for SUD, approximately 74\% had social media and 47\% reported their online communities to contain triggering content \cite{ashford2018technology}. Similarly, studies have shown up to 60\% of people felt they were treated unfairly due to having a SUD, and 39.5\% reported at least three types of stigmatizing experiences in their daily lives \cite{luoma2007investigation}. 

Despite widespread stigma in our society, there has been little agreement as to what constitutes stigmatizing language. Researchers have called for a standardized collection of terms or phrases \cite{kelly2004toward} which can be utilized to better serve as an assessment tool for the understanding and sensitivity regarding mental health, SUD, and its stigma. 
Additionally, the use of medically appropriate language by physicians and the general population can combat stigmatizing attitudes, offering respect for people with SUD~\cite{kelly2015stop}.

\subsubsection{Stigma and Hate Speech on Social Media}

The first tools used to counter stigmatizing and hateful posts online put the onus on social media users themselves to label posts as inappropriate \cite{kayes2015social}. Research efforts similarly utilized manually annotated data sets to label hateful and stigmatizing social media content \cite{golbeck2017large,mcneil2012epilepsy,founta2018large,davidson2017automated}. Studying hate speech in online text, particularly social media such as Facebook or Twitter \cite{macavaney2019hate}, has proved informative along the lines of gender \cite{waseem2016you,basile2019semeval}, religion \cite{albadi2018they}, race \cite{de2018hate, waseem2016hateful}, and immigration status \cite{basile2019semeval,ross2017measuring}. 

\begin{table*}[ht]
\resizebox{\textwidth}{!}{
\begin{tabular}{lc|lc|lc|lc|lc}\toprule
Keyword & N (\%) & Keyword & N (\%) & Keyword & N (\%) & Keyword & N (\%) & Keyword & N (\%) \\ \hline
acid & 379 (7.6) & dab & 77 (1.5) & lsd & 137 (2.7) & opiate & 127 (2.5) & shrooms &  73 (1.5)\\
adderall & 78 (1.6) & drug & 2396 (47.9) & marijuana & 174 (3.5) & opioid & 109 (2.2) & valium & 19 (0.4) \\
addy & 12 (0.2) & fentanyl & 35 (0.7) & mdma & 76 (1.5) & oxy & 24 (0.5) & weed & 762 (15.2) \\
cocaine & 128 (2.6) & heroin & 173 (3.5) & meth & 216 (4.3) & oxycodone & 5 (0.1) & xanax & 61 (1.2) \\
codeine & 18 (0.4) & kratom & 111 (2.2) & molly & 56 (1.1) & percocet & 4 (0.1) & xans & 16 (0.3) \\
coke & 242 (4.8) & kush & 28 (0.6) & norco & 3 (0.1) & purp & 7 (0.1) & xtc & 6 (0.1) \\ \bottomrule
\end{tabular}
}
\caption{Percentage of comments in the combined training and test data containing each substance keyword. 
}
\label{table:drug keywords}
\end{table*}

\subsubsection{Automatic Detection of Stigma}

Along with physical health conditions such as COVID-19 \cite{liu2022deep}, automatic stigma detection has recently been implemented in the context of mental health conditions such as depression, schizophrenia, and suicide \cite{li2018analysis,li2020comparison,jilka2022identifying,li2018detecting,oscar2017machine}. Automatic detection has also been applied to labeling social media content related to substance use \cite{roy2017automated,zhang2018utilizing}. However, stigma toward PWUS on social media is only beginning to be explored. 

 Perhaps closest to the present work, Chen et al. (2022)~\cite{chen2022examining} explore experiences of stigma, posted to the Reddit platform, by people who use substances. This work focuses on three types of stigma (anticipated, internalized, and enacted) and three substances (alcohol, cannabis, and opioids). While this paper also uses Reddit data and natural language processing techniques to understand stigma, it focuses on experiences of stigma as opposed to identifying stigmatizing content, which is the focus of the current study.

%% file: 03data.tex
\section{Data}
\label{sec: data}

\subsection{Annotation Data}


We begin with 1.66 billion Reddit comments from 2019 collected from pushshift.io~\cite{baumgartner2020pushshift}. We then identify comments which contain at least one substance use keyword (see below for keyword selection and disambiguation process) for a total of 9.3 million comments.
From this, we select 5,000 random comments for our annotation task. In Table \ref{table:drug keywords}, we break down the substance keyword distribution of these 5,000 comments. 

\paragraph{Substance Keywords} 
We identify comments related to substance use by identifying posts containing substance related keywords.  
These keywords were chosen to identify posts about specific substances (e.g., LSD or meth), a breadth of substances (e.g., we do not focus solely on opioids), general substances (e.g., drug*), and substance \emph{use} (e.g., smoke). 
The Drug Enforcement Administration slang word list was used as a starting point for choosing substance keywords~\cite{drug2018slang} and all keywords were agreed upon by an interdisciplinary team of substance use researchers. 
As a quality control check, we manually checked a random set of 1,000 posts in order to identify any obvious inconsistencies with our keyword data.
This was an iterative process where these manual check were discussed as a group and they keywords were further refined.
Through this process, we identified several simple heuristics designed to reduce false positives (i.e., comments that do not refer to substances) and, thus, removed comments containing the following phrases: hillary, clinton, obama, bernie, bern, sanders, trump, gab, weed out, crack jokes, crack me up, *white pill, black pill, red pill, blue pill, *whitepill*, *blackpill*, *redpill, *bluepill* and crazy pill. 
This resulted in 8,798,160 comments and is referred to as the Substance Keyword data set below.

\paragraph{Substance Keyword Disambiguation} We note that multiple keywords used to identify the Substance Keyword data set have multiple senses, many of which are not related to substances. For example, ``I smoked pot'' and ``I used a pot to cook.'' 
Thus, we attempt to refine our keyword list by removing keywords less likely to be referring to substances. To do this, two annotators were asked to rate 1,000 random comments (from the Substance Keyword data set above) for the following: ``Is this post about substances? Posts may reference substances by name, slang, or you may be able to determine by context.'' 
Both annotators are substance use researchers. 
The two raters agreed on 93.2\% of posts with a Cohen's Kappa score of 0.85. A total of 61.1\% of the 1,000 posts referred to substances (where both annotators agreed). 
Keywords were retained if they were used to discuss substances in more than 50\% of their occurrences or if they did not occur in the 1,000 random posts (using the assumption that these were rare words that would not dramatically increase false positives). 
This included: barbs, blunt, crack, ecstasy, joint, pot, and tabs.
The keywords ``acid'' was found to refer to substances in only 41\% of posts containing that keyword, yet, after internal discussions, it was decided that this keyword should be retained. 
The final list of keywords is shown in Table \ref{table:drug keywords}.

\begin{table}[!b]\centering
\begin{tabular}{lcc}\toprule
& Full Sample & Final Sample \\ \cmidrule(lr){2-2}\cmidrule(lr){3-3}
 & M (SD) or \# (\%) & M (SD) or \# (\%) \\ \hline
Age & 38.4 (10.5) & 38.8 (10.8) \\
Gender &   & \\
\hspace{3mm} Female & 241 (42.7\%) & 180 (45.0\%) \\
\hspace{3mm} Male & 371 (56.5\%) & 215 (53.8\%) \\
\hspace{3mm} Transgender, agender, etc. & 5 (0.80\%) & 5 (1.20\%)\\
Race / Ethnicity &   & \\
\hspace{3mm} African American & 80 (14.2\%) & 61 (10.8\%) \\
\hspace{3mm} Asian & 23 (4.1\%) & 20 (3.5\%) \\
\hspace{3mm} White & 447 (79.1\%) & 308 (77.0\%) \\
Substance Use &   & \\
\hspace{3mm} Know someone with a SUD & 382 (67.6\%) & 259 (64.8\%) \\
\hspace{3mm} Use substances & 369 (65.3\%) & 212 (37.5\%) \\
\hspace{3mm} Days of SU in the past 30 days & 6.6 (10.4) & 5.0 (9.7) \\ \bottomrule
\end{tabular}
\caption{Demographic distribution of MTurk workers. Full Sample $N=565$, Final Sample $N=400$.}
\label{tab:demographics}
\end{table}

\subsection{Model Evaluation Data}

\paragraph{Train and Test Split} The jury learning framework is an annotator-level model, which predicts each worker's annotations given comment text, past annotations, and group-level information. As such, we evaluate the model on unseen (or held out) Reddit \emph{comments}, as opposed to unseen workers or annotations. As described below in the Annotation Task section, the final annotated data set consists of 6,147 annotations of 3,802 comments from 400 workers. We create a random train/test split by taking 80\% and 20\% of the comments for training and testing, respectively. This results in 4,761 annotations across 3,042 comments in the training data and 1,386 annotations across 760 comments in the test data. This data set is used for \textbf{RQ1}.

\paragraph{Evaluating Stigma in the Wild} In order to identify stigmatizing language across Reddit (\textbf{RQ2} and \textbf{RQ3}), we apply the trained jury learning model to a large, random sample of unseen comments. As such, we collect 10,000 random comments from the Substance Keyword data set which do not appear in annotated data (i.e., do not appear in the training or testing data). 




%% file: 04annotationtask.tex
\section{Annotation Task}
\label{sec: annotation}

We begin by asking consenting Amazon Mechanical Turk (MTurk) workers a series of demographic questions (age, gender identity, and race/ethnicity), how many times they have used substances for non-medical reasons within the past 30 days, and if they know anyone who is in treatment for substance use disorder. 
Note that due to the potentially sensitive nature of these questions, we did not force a response to the survey, which has implications for the final sample (see below for final data filtering). 
We then give the workers a short training on the task, specifically what types of posts reference drugs (e.g., posts about drug stores or pharmacies should not be considered) and how to define stigma. 
Workers are then given a short three-question quiz, where they are shown the correct answer upon completion of the quiz (no workers are removed for incorrect quiz answers).  
See the Appendix for the full survey question text, quiz questions, and attention check. 

After completing the demographic survey and training materials, workers are then shown a series of 20 random Reddit comments. For each of the 20 comments, the workers are asked the question \emph{Q-Sub}: ``Are the drug-related words in this post being used to talk about drugs? (Yes/No)'' 
If the worker responded \emph{Yes}, then the second question \emph{Q-Stigma} is asked: ``Does this post contain stigmatizing language? (Yes/No)''.
This second question was skipped if the worker responded \emph{No} to \emph{Q-Sub}. 
For both questions, there is a 5-second delay before the submit button is displayed to force the worker to read the Reddit comment thoroughly. 
After 10 Reddit comments, an attention check question is asked (see Appendix for details on the attention check), which is immediately followed by the final 10 Reddit comments. 
The maximum number of annotations per comment was set to 3.

Workers are paid \$2.50 for completing the demographic questions and annotating 20 Reddit posts. 
In order to both view and work on this task, workers were required to be located in the U.S., have an approval rating of 80\% or higher, and have at least 100 approved HITs (Human Intelligence Tasks). 

\begin{figure}[!t]
\centering
\includegraphics[width=.5\columnwidth]{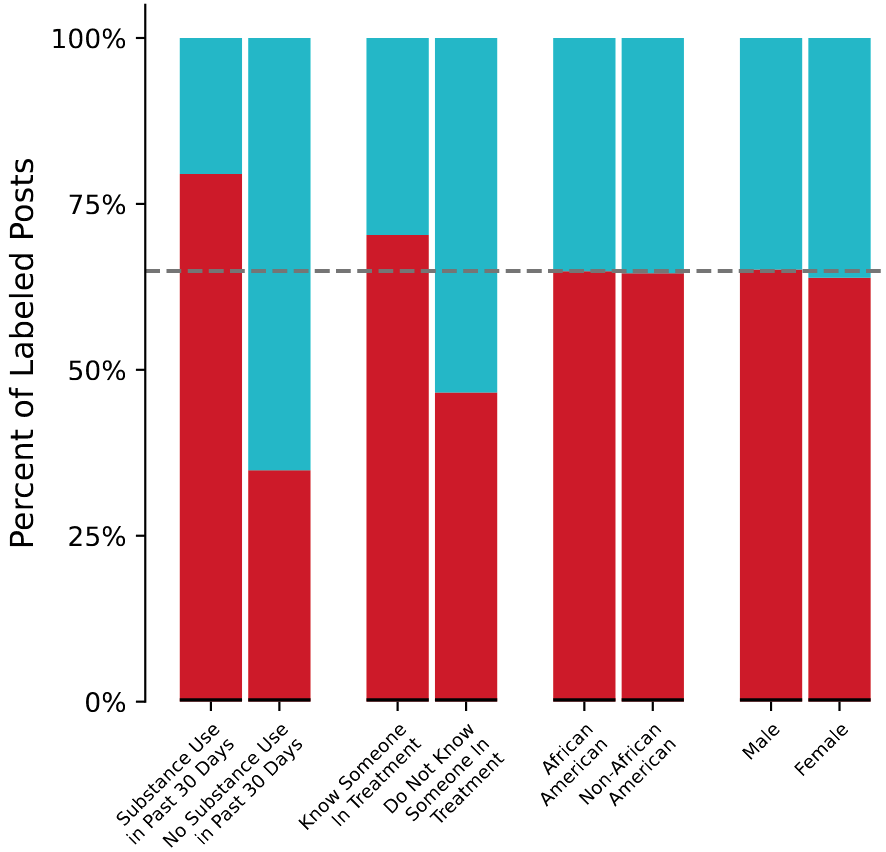}
\caption{Percentage of stigma labels (red) and non-stigma labels (blue) for each binary demographic group across the Full Sample. Grey dotted line is the percentage of positive stigma labels across the entire sample (65\%). }
\label{fig:labeled posts}
\end{figure}

\begin{figure}[!b]
\centering
\includegraphics[width=.5\columnwidth]{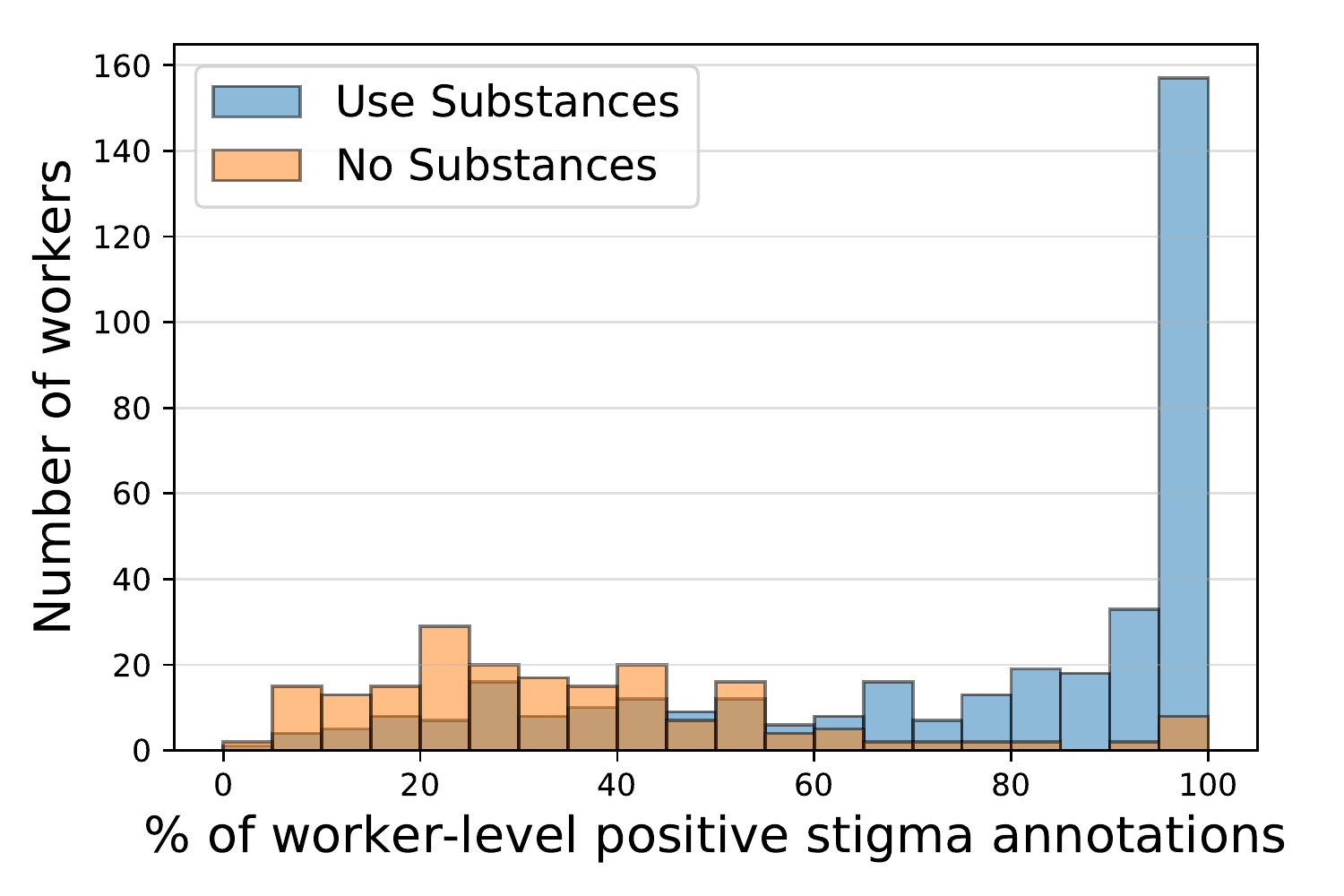}
\caption{Distribution of the worker-level percentage of positive stigma annotations  across the Full Sample (the number of positive stigma annotations divided by the worker's total number of annotations). PWUS are shown in blue and those who have not used substances are shown in orange. }
\label{fig:stigma histogram}
\end{figure}

\subsection{Annotation Results}
Due to the randomization process and workers not completing the full task, not all of the 5,000 comments were seen by workers, nor were all comments rated three times (our desired number of annotations per comment).
As such, at the end of the annotation process, a total of 4,991 comments were rated at least once by one of the 704 workers who attempted this task. 
We then removed workers who: (1) failed the attention check, (2) did not answer all questions in the demographic survey, and (3) did not complete the full series of 20 annotations in a single HIT (i.e., quit the task early). 
Note that if the worker answered \emph{No} to \emph{Q-Sub} then they were not asked to rate the comment for stigma. 
Thus, we further refined the data set to only those comments which were annotated for stigma at most three times. 
This produced a data set of 4,600 comments rated at least once for stigma by 565 workers for a total of 9,392 annotations. 
The demographic distribution of this worker sample is seen in the Full Sample column of Table \ref{tab:demographics}.

Next, we examined the distribution of positive stigma labels (i.e., stigma is present in the comment) across the workers. In Figure \ref{fig:labeled posts}, we see the percentage of labeled posts across each of the different demographic groups: those who use substances, those who know someone in treatment for a SUD, gender, and race/ethnicity. We see significant differences (via two-sided t-test) in the number of positive stigma labels across those who use/do not use substances ($t=48.4$, $p<0.01$) and those who know/don't know someone in treatment ($t=16.4$, $p<0.01$). Gender ($t=0.62$, $p>0.05$) and race/ethnicity ($t=-0.58$, $p>0.05$) differences are not statistically different. In Figure \ref{fig:stigma histogram}, we plot a distribution of the percentage of positive stigma labels across each worker's total annotations (the number of positive stigma labels divided by the worker's total number of annotations). We plot this distribution for both people who have used substances at least once in the past 30 days (blue) and those who did not (orange). As seen in this figure, there is a large spike at the tail end where workers rated over 95\% of their annotations as stigmatizing. Notably, the majority of people in this bin use substances. While this behavior may point to bad data (random or unreliable annotations), we note that the workers in this bin passed the attention check, fully responded to the demographic survey, and did not always positively identify each post as referring to substances (\emph{Q-Sub}). Additionally, if this was a sign of unreliable workers, it should be distributed randomly across the demographics. However, this pattern only holds across those who use substances and those who know someone with a substance use disorder and not age, gender, and race/ethnicity. 

While we believe these annotations to be useful data, they cause an imbalance in our data set. 65\% of the annotations are labeled as stigmatizing, which is much higher than previous studies on toxic language, which tends to find positive labels rare. Therefore, we removed 165 workers who rated at least 95\% of their annotations as positive for stigma. This leaves a final data set of 6,147 annotations across 3,802 comments from 400 workers. Their demographic distribution is found in the Final Sample column of Table \ref{tab:demographics}. This final data set has 47\% of annotations labeled as stigmatizing. 

In Table \ref{tab:group agreement} shows the agreement (Krippendorff's $\alpha$) for each subgroup (e.g., gender and substance users). The agreement across all groups is $\alpha=0.12$. We see small differences between Female ($\alpha=0.14$) and Not Female ($\alpha=0.13$), as well as African Americans ($\alpha=0.16$) and Not African Americans ($\alpha=0.13$). On the other hand, we see larger differences between those with lived experience with substance use. Taken with the results above, we see that those with lived experience are more likely to label a post as stigmatizing but also do not agree on which posts are stigmatizing. 

\begin{table}[]
\centering
\begin{tabular}{lc}
\toprule
 &  Krippendorff's $\alpha$ \\ \midrule 
All                               & .12 \\ 
\emph{Gender} & \\
\hspace{3mm} Female                               & .14 \\
\hspace{3mm} Not Female                           & .13 \\
\emph{Race / Ethnicity} & \\
\hspace{3mm} African American                     & .16 \\
\hspace{3mm} Not African American                 & .13 \\
\emph{Knowing Someone in Treatment} & \\
\hspace{3mm} Know Someone             & .12 \\
\hspace{3mm} Do Not Know Someone     & .18 \\
\emph{Substance Use in the Past 30 Days} & \\
\hspace{3mm} Substance Use     & .09 \\ 
\hspace{3mm} No Substance Use  & .31 \\ 
\bottomrule
\end{tabular}
\caption{Krippendorff's $\alpha$ for each subgroup.}
\label{tab:group agreement}
\end{table}

%% file: 05methods.tex
\begin{table}[!b]
\centering
\begin{tabular}{clccc}\toprule
 &  & Accuracy & F1 & AUC \\ \hline
\parbox[t]{2mm}{\multirow{5}{*}{\rotatebox[origin=c]{90}{Baselines}}}  & Most Frequent Class & .50 & .33 & .50 \\ 
 & LIWC$^\dagger$ & .55 & .54 & .59 \\ 
 & LIWC + Dem.$^\dagger$ & .59 & .58 & .63 \\ 
 & Unigrams$^\ddagger$ & .58 & .57 & .61 \\  
 & Unigrams + Dem.$^\ddagger$ & .63 & .63 & .67 \\ 
 \hline
\parbox[t]{2mm}{\multirow{3}{*}{\rotatebox[origin=c]{90}{DCN}}} & BERTweet & .59 & .58 & .59 \\
 & BERTweet + Dem. & .64 & .64 & .64 \\
 & BERTweet + Dem. + Ann. & \textbf{.69} & \textbf{.69} & \textbf{.69} \\ \bottomrule
\end{tabular}
\caption{Annotation level predictive accuracy. $\dagger$ logistic regression, $\ddagger$ extra trees classifier.}
\label{tab:annotator level results}
\end{table}

\section{Methods}
\label{sec: methods}

Our analysis proceeds in three steps: (\textbf{RQ1}) training and evaluating the jury learning model, (\textbf{RQ2}) evaluating the effect of demographic representation within the jury, and (\textbf{RQ3}) identifying linguistic cues associated with stigma. First, we train and evaluate an annotator-level jury learning model to assess how well our model can classify stigma annotations from text, person-level information, and group-level information. Next, we apply the trained jury model to a set of unseen Reddit comments and assess how different jury configurations (e.g., juries consisting of PWUS and those who do not) change the final stigma label. Finally, we identify language associated with labels and examine where different populations agree/disagree on stigmatizing content in order to understand how people perceive stigma.

\begin{figure*}[!h]
\centering
\includegraphics[width=.9\textwidth]{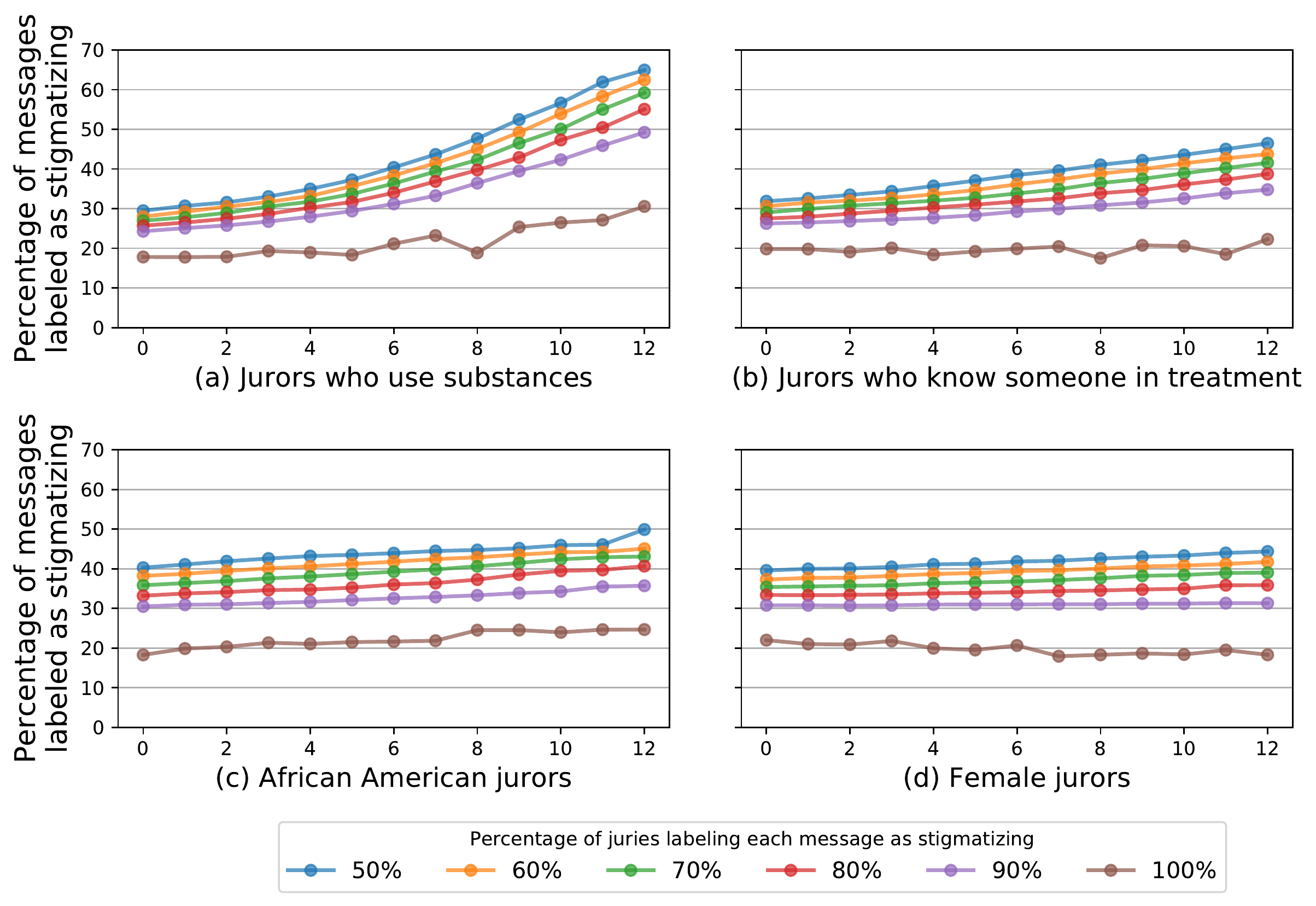}
\caption{The effect of the jury demographic distribution on the amount of stigmatizing content across the data set. Each dot is the percentage of the 10,000 comments labeled as stigmatizing, where each comment is rated by 10,000 random juries. The color of the line indicates the threshold used to assign the final comment label from the 10,000 jury votes (e.g., a blue line means that a comment is labeled as stigmatizing if at least 50\% of the 10,000 juries vote positive stigma). Within each plot, moving from left to right increases the representation of jurors with a given demographic within each of the 10,000 juries (e.g., juries in (d) at 0 are all male and at 12 are all female).}
\label{fig:distributions}
\end{figure*}

\subsection{Jury Learning}
Jury learning is a supervised machine learning framework used to predict an individual annotator's label on unseen examples (in our case, stigmatizing content on Reddit) developed by Gordon et al. (2022)~\cite{gordon-jury-learning}. Jury learning draws on the notion of juries in the U.S. legal system, specifically through the use of group voting (as opposed to single judge) and the jury selection process. Modeling each annotator in the data set grants practitioners the ability to define the representation of groups of people in the training data. Thus, practitioners can build a jury relevant to the problem at hand from a large pool of annotators. In our example, since the final desired label is whether or not a Reddit post is stigmatizing towards PWUS, we might want the majority of our jurors to know someone in treatment for a SUD. It is easy to imagine similar examples for other toxicity detection-related tasks, such as annotating gender stereotypes, hate speech towards racial minorities, etc. 

The learning architecture follows a Deep and Cross Network (DCN), which is a standard recommender system architecture~\cite{wang2021dcn}. Using a movie recommender system as an example, the recommender system will model the movie itself, a person's past viewing history, and who that person is. In more general terms, this architecture models three distinct pieces of information: content (i.e., the movie), group (i.e., who this person is), and person (i.e., the person's viewing history). Applied to this setting, the DCN models the Reddit comment (content), worker demographics (group), and each worker's annotation history (person). 

More formally, the DCN consists of an embedding layer, a cross-network, and a deep network. The embedding layer is a concatenation of the content, group, and person embeddings. This is then fed into the cross-network, which models the explicit interactions between the three embedding types through the use of cross layers. The output of the cross-network is then fed into the deep network (a standard feed-forward network) to model implicit interactions. 

We emphasize the fact that the end goal for the DCN is a trained architecture that can be used to model each worker in the annotation task (\textbf{RQ1}). Thus, the DNC is used to model both the annotators and the data (i.e., the Reddit posts). We further emphasize the fact that during the training process we have not introduced the idea of the ``jury" (i.e., jury learning is not limited to DCNs). This is done after we have trained the DCN and applied the model to unseen data to generate stigma predictions from each annotator (\textbf{RQ2} and \textbf{RQ3}).

\subsection{RQ1: Automatically Identifying Stigma}

We initialize our model and vectorizer with a standard BERTweet model~\cite{bertweet,liu2019roberta}. Using a similar training procedure to Gordon et al. (2022)~\cite{gordon-jury-learning}, we train the BERTweet model and the DCN on five epochs of our data, allowing it to alter the underlying BERTweet model. We then freeze the BERTweet model and train for 15 more epochs. We note that Gordon et al. do an initial fine-tuning step where the BERTweet model is fine-tuned on a large toxicity data set. We chose not to do this initial fine-tuning as the data set is focused on toxicity detection, which may be different from substance use-related stigma. Since our end goal is to understand stigmatizing content, we felt that this fine-tuning process might introduce unintended biases toward general toxicity. 

As described in the Data section, the DCN is trained on 80\% of the comments in the final annotation data and evaluated on the remaining 20\% of comments. We use BERTweet's pooler output as the content embeddings. We include five worker-level demographics (group embedding): continuous age, binary gender (1 for female, 0 otherwise), binary race/ethnicity (1 if African American, 0 otherwise)\footnote{Our analysis includes very narrow senses of gender and race/ethnicity and we do not mean to imply that either construct is binary.}, knowing someone in treatment (1 if the worker knows someone, 0 otherwise), and the (continuous) number of times the worker used substances in the past 30 days. We also include a one-hot encoding of each worker (person embedding). See Appendix for full details and hyperparameter values. Out-of-sample classification accuracy is measured via accuracy, F1, and Area under the ROC Curve (AUC).

\subsubsection{Baselines}
We compare the DCN to two baselines: Linguistic Inquiry and Word Count (LIWC) and unigrams. In all baselines, we consider models trained on (1) text-based features and (2) text-based features plus worker-level demographics (age, gender, race/ethnicity, substance use in the past 30 days, and knowing someone in treatment for a SUD). We also include a simple Most Frequent Class (MFC) classifier. All features are used within either a logistic regression (LIWC) or an Extra Trees Classifier (unigrams). We use the same training and test data as the DCN. See the Appendix for full details and hyperparameter values (which are set via 10-fold cross-validation on the training data). 

\paragraph{LIWC} LIWC is a dictionary consisting of 73 manually curated categories, including both content and function words, and is one of the most widely used dictionaries in social and psychological sciences~\cite{pennebaker2015development}. Example categories include positive and negative emotions, pronouns, verbs, and adjectives. 

\paragraph{Unigrams} We extract unigrams using a tokenizer designed for social media data~\cite{dlatk}. On the training dataset, there are 17,752 unique unigrams. In order to keep the number of features less than the number of observations in the training data (4,761 annotations), we remove rare unigrams: any unigram used by less than .5\% of the training data (24 annotations). This results in a total of 1,312 unigrams in the final feature space.

\subsection{RQ2: Effects of Jury Representation}

Here we attempt to answer \textbf{RQ2}: Does lived experience with substance use inform how often stigma is perceived? We begin with four binary demographic splits: female/not female, African American/not African American, uses substances/does not uses substances, knows/does not know someone in treatment. Given a fixed jury size of 12, we consider all possible jury configurations for each of the four demographic splits. For example, a jury with 0 female/12 non-females, 1 female/11 non-females, etc. Then for each of the 10,000 comments in the evaluation data, we create 10,000 random (without replacement) juries for the given configuration (e.g., 10,000 juries with 1 female and 11 non-females). Next, for each of the 10,000 juries, we apply the trained jury model to the jurors to produce 12 stigma ratings (corresponding to the 12 jurors) for the comment. We assign a stigma label of 1 if more than half (7) of the jurors vote that the comment is stigmatizing, and assign 0 otherwise. Thus, for each of the 10,000 comments, we have 10,000 labels, each label produced from the majority vote of the random juries. 

Next, in order to assign a final stigma label from the 10,000 jury ratings, we consider increasingly stricter thresholds on the percentage of positive stigma votes needed to assign the final label. For example, we begin by assigning a final label to a comment if 50\% of the 10,000 juries rate the comment as stigmatizing and increase this threshold up to 100\%. We do this for each of the 10,000 comments in the evaluation data set and look at the total percentage of stigmatizing posts across the entire data in order to see if this percentage changes as the demographic representations across the juries change.

\subsection{RQ3: Stigmatizing Language in the Wild}
In order to identify stigmatizing language (\textbf{RQ3}), we first assign labels to each of the 10,000 Reddit comments using two separate juries types: a jury where all 12 members have used substances within the last 30 days and a jury where all 12 members have no used substances. At the individual jury level, a stigma vote is assigned if at least 7 members (i.e., the majority) vote that the comment is stigmatizing. As described above, we select 10,000 random juries (for each jury type) and label a comment as stigmatizing if at least 90\% of juries vote that the comment is stigmatizing. Thus, in the end, we have two labels for each of the 10,000 comments: one label each from the two jury types.

For the first step of this analysis, we want to know where all juries \textbf{Agree} on stigma. Thus, we only consider comments where both jury types agree (e.g., substance-using juries vote \emph{yes} stigma and non-substance-using juries also vote \emph{yes} stigma). We then examine language features (LIWC and unigrams) associated with the binary stigma label. To do this, we perform a single regression for each feature in our feature space (a process called Differential Language Analysis or DLA \cite{schwartz2013personality}). In particular, for each language feature, we perform a logistic regression where the independent variable is the relative frequency of a given language feature and the dependent variable is a binary variable set to 0 where both jury types vote \emph{no} stigma and 1 where both jury types vote \emph{yes} stigma. The LIWC category and unigram frequencies are standardized (mean-centered and divided by the standard deviation). Due to the large number of comparisons, we perform a Benjamini-Hochberg False Discovery Rate (FDR) correction and only consider associations significant at a corrected rate of $p<0.05$~\cite{benjamini1995controlling}. 

For the second step of this analysis, we want to know where people with lived experience with substance use see stigma, but those who do not have the same lived experience do not see stigma. Thus, we only consider comments where substance-using juries vote \emph{yes} stigma and examine where non-substance-using juries \textbf{Disagree}. Again, we perform a series of independent logistic regressions (i.e., DLA) using LIWC category and unigram frequencies as the independent variables and a binary dependent variable: 0 where non-substance using juries vote \emph{no} stigma and 1 where non-substance using juries vote \emph{yes} stigma. Again, LIWC category and unigram frequencies are standardized and we apply a Benjamini-Hochberg FDR correction. In both steps, effect sizes are reported as a Cohen's D: the mean difference between the two groups (the 0 and 1 binary labels) divided by the pooled standard deviation.

%% file: 06results.tex
\section{Results}
\label{sec: Results}

Table \ref{tab:annotator level results} shows the results of the jury learning process (\textbf{RQ1}). Here we see that across all models, adding demographic features increases the predictive accuracy over Reddit comment language alone. We also see that using more sophisticated language features (e.g., BERTweet vs LIWC) increases predictive accuracy. In the end, the DCN (jury learning) using all three feature types (content, person, and group) outperformed all other models. Thus, we can answer \textbf{RQ1} in the affirmative: stigma towards PWUS can be automatically identified via machine learning methods.

In Figure \ref{fig:distributions}, we see the results of the jury learning model applied to 10,000 Reddit comments (\textbf{RQ2}). Juries with lived experience with substance use (either those who use substances or those who know someone in treatment) tend to label more content as stigmatizing and this increases as their representation within each jury increases. On the other hand, we do not see such pronounced increases across gender or race/ethnicity, both of which are marginalized populations and could be sympathetic to stigma (and thus see stigma where others may not). Here we see slight increases as both African Americans (Figure \ref{fig:distributions}(c)) and females (Figure \ref{fig:distributions}(d)) represent a larger portion of each jury. In reference to \textbf{RQ2}, we see that lived experience with substance use increases the \emph{frequency} at which people perceive stigma, and this is not true for other groups. 

\begin{figure}[t]
\centering
\includegraphics[width=.5\columnwidth]{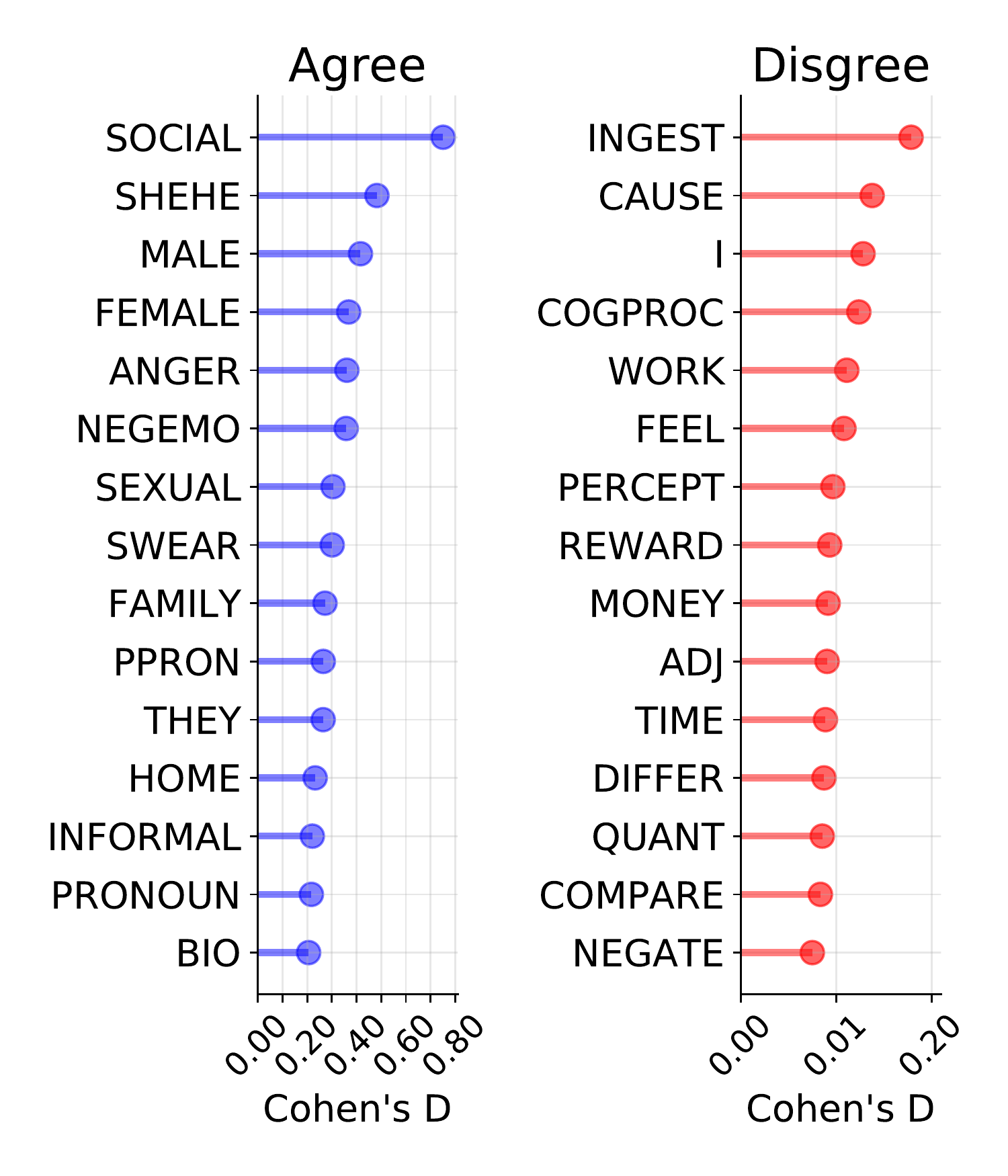}
\caption{LIWC categories associated with the presence of stigma in comments where both substance using and non-substance using juries agree and disagree (substance using juries labeled as stigma and non-substance using juries did not). All correlations significant at a Benjamini-Hochberg significance level of $p < 0.05$. Note the x-scales of the two plots are different.}
\label{fig:liwc correlations}
\end{figure}

\begin{figure}[t]
\centering
\includegraphics[width=.5\columnwidth]{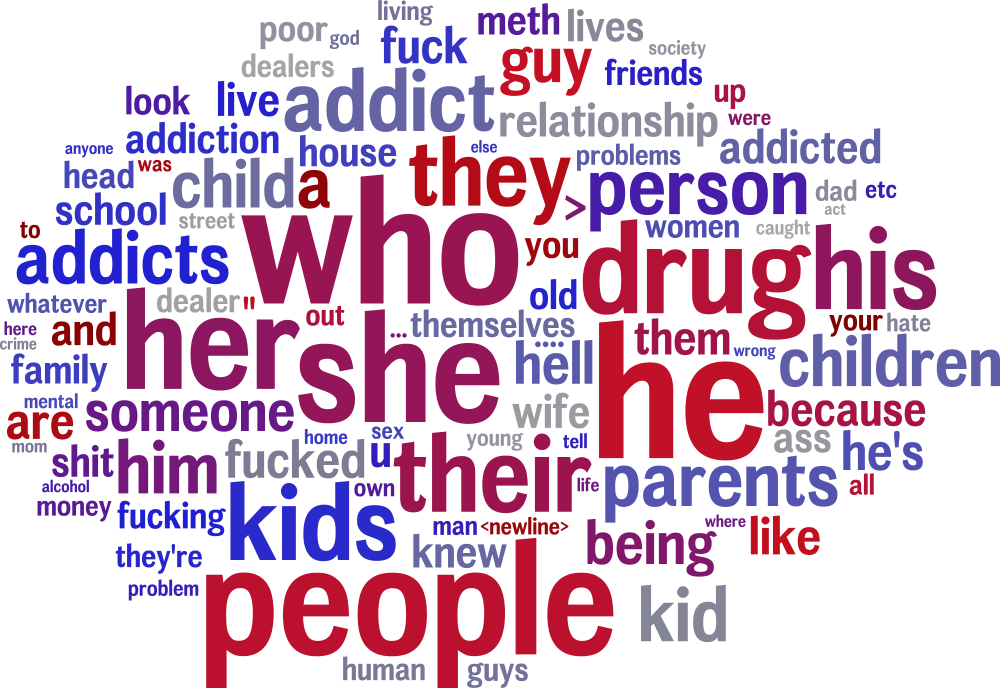}
\caption{Unigram's associated with the presence of stigma in comments where both substance using and non-substance using juries agree. Size of the word indicates the strength of the correlation (larger Cohen's D); color indicates relative frequency of usage (rare words are grey,  frequent words are red, with blue in-between). All correlations significant at a Benjamini-Hochberg significance level of $p < 0.05$. Correlations (Cohen's D) range from 0.07 to 0.31. 
}
\label{fig:unigram correlations}
\end{figure}

In Figures \ref{fig:liwc correlations} and \ref{fig:unigram correlations}, we see language correlated with stigmatizing content (\textbf{RQ3}). In Figure \ref{fig:liwc correlations}, the \textbf{Agree} label is where both jury types (PWUS and those who do not) agree on stigma/no-stigma, while the \textbf{Disagree} label is where juries with PWUS see stigma, but those who do not use substances do not see stigma. As seen in Figure \ref{fig:liwc correlations} (\textbf{Agree} only), stigma is associated with the ANGER, SWEARING, NEGEMO (negative emotions), and SEXUAL categories. The SHEHE category is 3rd person singular pronouns, while PPRON and PRONOUN are general pronoun categories. 

Figure \ref{fig:unigram correlations} gives further context to these results. Here we see references to others (``people'', ``he'', ``she'', ``they'', ``their''), ``addicts'' and ``addiction'', ``dealers'', and references to children, parents, and schools. Notably, only a single substance is mentioned ``meth'', which is a highly stigmatized substance~\cite{deen2021stigma} and the focus of dehumanizing protroyals in the media and in anti-drug campaigns~\cite{habib2023role}.

LIWC correlations with the \textbf{Disagree} label include INGEST, COGPROC (cognitive processes), and I (first person singular pronouns). Only a few unigrams were associated with the \textbf{Disagree} label: ``acid'', ``coke'', ``i'', ``weed'', and ``the''. Notably, this includes three specific substance types, whereas the \textbf{Agree} results in \ref{fig:unigram correlations} do not contain many references to substances, other than ``meth'' which is generally found to be associated with stigmatizing or dehumanizing content~\cite{linnemann2013your}. 

While the stigmatizing words associated with \textbf{Agree} contain mentions of others, the LIWC category I and the unigram ``i'' are both associated with \textbf{Disgree}. Thus, juries which contain PWUS identify stigma in comments which contain self-references, where juries who do not use substances do not find this. 

Summarizing these results, in order to answer \textbf{RQ3}, we see both similarities and differences between how stigma is perceived between those with lived experience and those without. These two groups agree that negative emotions, swearing, and outgroups (or othering) is indicative of stigma. They also disagree with self-focus and mentions of substances being more stigmatizing for those with lived experience.

%% file: 07discussion.tex
\section{Discussion}
\label{sec: discussion}

The results show that (1) those with lived experience perceive more stigma (Figures \ref{fig:labeled posts} and \ref{fig:stigma histogram}), (2) those with lived experience do not always agree on what is stigmatizing, and (3) there is overlap between those with lived experience and those without in what words are more like to be stigmatized (Figures \ref{fig:liwc correlations} and \ref{fig:unigram correlations}). Taken together, there is some agreement across groups on what is stigmatizing, but it is a highly subjective and personal experience. Thus, we believe there are similar takeaways as those from the hate speech/toxicity literature: (1) we need diverse views, (2) we need to include and center those with lived experience, and (3) there is not a ``one size fits all'' approach to dealing with stigma.

The results also suggest that some substances (``meth'') may be more stigmatized than others, at least on social media platforms. This differential stigmatization of substances is important to consider given the current political situation in the U.S., where many states are legalizing marijuana while simultaneously defunding harm reduction techniques (such as safe-use sites).  

The linguistic correlates of stigma (\textbf{RQ3}) dovetail with those found by Chen et al. (2022)~\cite{chen2022examining}. This study also found mentions of family, swearing, negative feelings, and self-focus. Notably missing from our results are mentions of recovery, quitting, and withdrawal, which were all found by Chen et al., which might result from their focus on posts about stigmatizing experiences rather than stigmatizing content. 

One potential limitation is that the substance use questions and the definition of stigma were presented \emph{before} the actual annotation task. Thus, the annotations could be influenced by the demographic questions and the training material (i.e., priming). We chose to collect the demographic data before the annotation task to maximize the data collected (i.e., if someone does not complete the demographic questions, their annotation data is dropped). This was done since the priming effect could go both ways: substance use questions could influence the annotations and the annotation task could influence the answers to the substance use questions. Similarly, our definition and examples of stigmatizing content (used in the training process before annotations were collected) could also influence how the workers annotated the posts. To minimize this influence, we used examples that we felt were extreme and, thus, likely to be seen as stigmatizing by most people. 


%% file: 08conclusions.tex
\section{Conclusions}
\label{sec: conclusions}

Our findings show that people with lived experience with substance use perceive more stigma and respond to different linguistic signals than other marginalized or minority groups (gender or race/ethnicity). Despite this, there is consensus on what types of language lead to stigmatizing content, for example, othering, swearing, and terms like ``addict'' (though by no means do we wish to imply that stigma is universally agreed upon). 

Having a jury-based process for determining stigmatizing language provides a way to gain insights from data by centering diverse groups of workers. This process allows us to conduct a deeper linguistic analysis of messages where those with lived experience with substance use and those who do not differ in their determination of stigma. Our hope is that analysis will reveal concepts and language that seem acceptable to those creating messaging for the substance use audience but are actually harming the recipient. Conversely, there may be messaging that would seem harmful but would instead be impactful.

 
\subsection{Broader Perspective}

There is increased attention to stigma in both mental health and substance use. This has become especially important in recent years as the U.S. deals with the opioid epidemic, with approximately 1 million deaths since 1999~\cite{hedegaard2022drug}. As such, there have been several calls for clinicians and researchers to choose the words they use to describe these stigmatized populations~\cite{volkow2021choosing}. For example, using the term ``poisoning'' instead of ``overdose'' to avoid implications that (1) there is a correct and safe dose and that (2) the substance user knows what a proper dose is and chooses to take more. Similarly, NIDA, previously the National Institute on Drug Abuse, has changed its name to the National Institute on Diseases of Addiction. Thus, a linguistic analysis of stigma towards PWUS, which centers around people with lived experience, is timely and may help reveal where additional work is needed~\cite{stull2022potential}.

\subsection{Ethical Considerations}

The study protocol was approved by the Institutional Review Board at REDACTED. MTurkers were consented and informed on the nature of the study and what data was being collected. As with all studies using public social media data, none of the Redditors consented to have their comments collected, analyzed, and publicly released. See Chancellor et al. (2019)~\cite{chancellor2019human} for a thorough discussion of the ethical issues (including informed consent) of ``humans'' in ``human-centered machine learning.'' 

We have taken care to anonymize the released data. This includes (1) hashed Mechanical Turk worker-ids, (2) a manual check for identifying information within the Reddit comments, and (3) only releasing Reddit comment-ids instead of releasing the entire comment text. Only releasing comment-ids allows Redditors to delete their posts from the platform, preventing any future data pulls from collecting their labeled posts. Following the recommendations in Gebru et al. (2021)~\cite{gebru2021datasheets}, our released data set includes a data sheet with information related to motivation, funding, the collection process, etc. 

The Mechanical Turk HITs were written in monolingual English and workers were required to live in the U.S. Similarly, the LIWC dictionaries and BERTweet model used in this study only contain monolingual English. Therefore, the results presented here may not generalize outside the U.S. or to non-English languages or minority populations.

%% file: supplement.tex
\section{Annotation Task Details}


\paragraph{Demographics Survey} We ask the following demographic questions: (1) What is your age (in years)?; (2) What is your gender identity? Mark all that apply: Male, Female, Genderqueer, Agender, Transgender, Cisgender, A gender not listed (open ended); and (3) With which racial and ethnic group(s) do you identify: American Indian/Alaska Native, Hispanic, Latino/a, or Spanish origin, White, Asian, Middle Eastern or North African, Black or African American, Native Hawaiian or Other Pacific Islander, Another race or ethnicity not listed above (open ended). 

\paragraph{Substance Use Related Questions} We ask the following questions related to substance use: (1) How many times in the past 30 days have you used an illegal drug or used a prescription medication for nonmedical reasons? For instance, because of the experience or feeling it caused; and (2) Do you know anyone close to you who has ever had treatment (including addiction medication) or attended support groups (ex. AA, NA) for substance or alcohol use disorder: Yes/No. 

\paragraph{Quiz} We ask workers to answer the following three question quiz: (1) What should NOT be considered drug-related: mentions of a drug store, Insurance plans and prescription coverage, Posts written in a language other than English, All of the above; (2) If you saw the following post, how would you rate it? ``I have been taking extra medications because I like the way it makes me feel'': Yes, contains drug terms / No, does not contain drug terms; and (3) Which of the following might be considered stigma, or stigmatizing language? Mentions of: prejudice, discrimination, stereotypes, all of the above, none of the above. 

\paragraph{Attention Check} The following fake Reddit comment was created to be both unambiguously about substances and unambiguously stigmatizing: ``The average burglar is just some idiot, 20 year old crackhead. Stealing from people like me to buy drugs for them and their addict friends. They get what they deserve and belong in jail. At least they can't collect unemployment there.''
Workers were then asked the same two questions as all other Reddit comments: ``Are the drug-related words in this post being used to talk about drugs?'' and ``Does this post contain stigmatizing language?''.
This attention check was asked at the halfway point in each task (after 10 of the 20 annotations).
Given an incorrect answer (\emph{No} to either of the two questions), the associated annotations were deleted. 
Despite the fact that the data were not used in the final data set, workers were able to complete the task and were compensated the full amount.

\section{Experimental Parameters}
Following Gordon et al.~\cite{gordon-jury-learning}, we use a DCN of 3 cross layers followed by 3 deep layers of size 768 (standard Multi-Layer Perceptrons with ReLU activation) finally feeding into a logit layer with an output of a single real number. We use an Adam optimizer with learning rate = $1\times 10^{-5}$. Models are trained are trained using NVIDIA RTX A6000 GPU.

The baseline models include a logistic regression and extra trees classifier (ETC). Logistic regression parameters: C = 1000000 (for an $l_0$ penalty approximation), penalty = l2, dual = False, and random\_state = 42. ETC parameters: n\_jobs = 12, n\_estimators = 1000, max\_features = sqrt, criterion = gini, min\_samples\_split = 2, class\_weight = balanced\_subsample. Unless otherwise specified, all default values are used. Each classifier is implemented using the scikit-learn Python package~\cite{scikit-learn} within the DLATK Python package~\cite{dlatk}.